\pdfoutput=1
\documentclass[11pt]{article}
\usepackage[final]{acl}

\usepackage{times}
\usepackage{latexsym}

\usepackage[T1]{fontenc}

\usepackage[utf8]{inputenc}

\usepackage{microtype}

\usepackage{inconsolata}

\usepackage{graphicx}

\usepackage{times}
\usepackage{soul}
\usepackage{url}
\usepackage{amsmath}
\usepackage{amsthm}
\usepackage{booktabs}
\usepackage{algorithm}
\usepackage{algorithmic}
\usepackage[switch]{lineno}
\usepackage{stfloats}
\usepackage{placeins}

\usepackage{makecell}
\usepackage{cleveref}

\usepackage{url}
\usepackage{hyperref}
\usepackage{color}
\usepackage{amssymb}
\usepackage{mathabx}
\usepackage{mdframed}
\newcommand\cnot[1]{%
  \mathrel{\ooalign{\hfil$#1$\hfil\cr\hfil$/$\hfil\cr}}}

\DeclareMathOperator*{\argmin}{arg\,min}

\newcommand{\token}[1]{\textless #1\textgreater}

\title{Out-of-Context Reasoning in Large Language Models}

\author{
 \textbf{Jonathan Shaki\textsuperscript{1}},
 \textbf{Emanuele La Malfa\textsuperscript{2}},
 \textbf{Michael Wooldridge\textsuperscript{2}},
 \textbf{Sarit Kraus\textsuperscript{1}},
\\
\\
 \textsuperscript{1}Bar-Ilan University,
 \textsuperscript{2}University of Oxford,
}

\begin{document}
\maketitle

\begin{abstract}

We study how large language models (LLMs) reason about memorized knowledge through simple binary relations such as equality ($=$), inequality ($<$), and inclusion ($\subset$). Unlike in-context reasoning, the axioms (e.g., $a < b, b < c$) are only seen during training and not provided in the task prompt (e.g., evaluating $a < c$). The tasks require one or more reasoning steps, and data aggregation from one or more sources, showing performance change with task complexity. We introduce a lightweight technique, out-of-context representation learning, which trains only new token embeddings on axioms and evaluates them on unseen tasks. Across reflexivity, symmetry, and transitivity tests, LLMs mostly perform statistically significant better than chance, making the correct answer extractable when testing multiple phrasing variations, but still fall short of consistent reasoning on every single query. Analysis shows that the learned embeddings are organized in structured ways, suggesting real relational understanding. Surprisingly, it also indicates that the core reasoning happens during the training, not inference.
\end{abstract}

\section{Introduction}
\begin{figure*}
    \centering
    \includegraphics[width=1\textwidth]{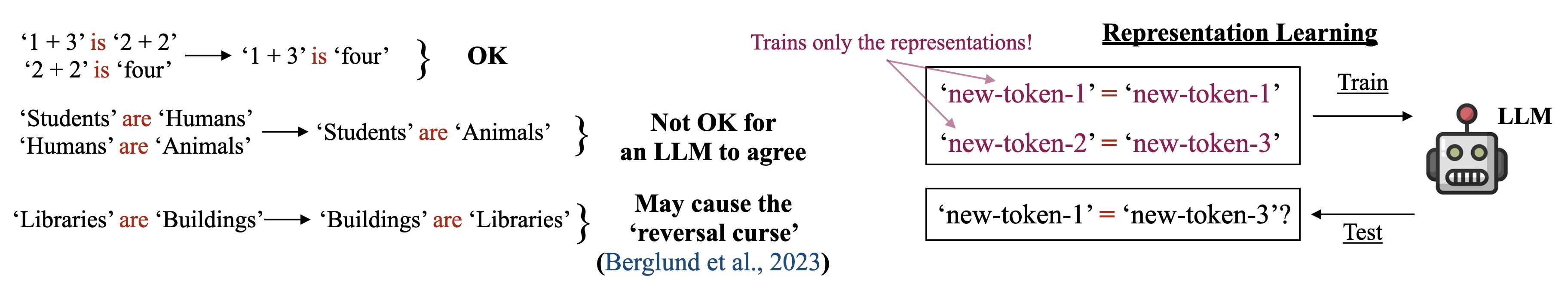}
    \caption{Left: an example of a binary relation that a model would learn without issues (top). On the other hand, both the examples at the bottom raise issues in terms of the acceptability of the answer (which nonetheless would follow from an axiomatic system). Right: our solution encompasses all three cases without falling into the biases of existing representations.}
    \label{fig:intro-illustration}
\end{figure*}
A large number of works have investigated the reasoning capabilities of Large Language Models (LLM), spanning from math~\cite{frieder2023mathematicalcapabilitieschatgpt}, logic~\cite{kojima2023largelanguagemodelszeroshot,pan2023logic}, planning~\cite{guan2023leveraging,lin2024graphenhancedlargelanguagemodels,valmeekam2024planbench}, and, more recently, multi-agent problem solving~\cite{li2024agentsneed}.
The empirical evidence suggests that the larger a model and its associated training data, the more capable the LLM is at handling \textit{unseen} problems~\cite{brown2020languagemodelsfewshotlearners,kaplan2020scaling}.
Complex problem-solving relies on the capabilities of a model to decompose a problem into its sub-components and, similarly to a puzzle, provide the correct answer by integrating the solutions from each sub-task.
This principle, known as \emph{compositionality}~\cite{dziri2023faithfatelimitstransformers}, relies on the assumption that LLMs possess a core set of capabilities to solve each sub-task with low error probability.  
Most existing benchmarks focus on in-context reasoning, where the necessary information is explicitly provided within the prompt~\cite{mccoy2023embersautoregressionunderstandinglarge}. This approach offers insights into how models process information presented at inference time. Yet, in-context learning does not assess the ability of LLMs to reason out-of-context, i.e., based on memorised knowledge encountered only during training and that does not appear in the prompt.
\newline \newline
While several works have explored out-of-context learning~\cite{allen2023physics,hu2024limited,zhu2024beyond}, they primarily focus on complex/high-order tasks, making it hard to identify the reasons behind a model's success or failure.
The most closely related work to ours is~\cite{berglund2023reversal}, where the authors explored LLMs' difficulty with reversal relations, summarised by its title "LLMs trained on `A is B' fail to learn `B is A'". The verb ``is'' can be interpreted as a binary relation or as a verb, introducing a confounder that makes it complex to judge whether a model captures the core properties of transitivity. While in logic, ``A is B'' implies ``B is A'', from a linguistic perspective, ``Students are Humans'' poses some issues in deriving that "Humans are Students", and it is thus hard to impute a model's failure to its inability to handle symmetry properly. In addition, they tested whether a model generates A given B, instead of whether ``B is A'' evaluates true, implicitly assuming that only high-probability predictions are those that a model considers correct. Figure~\ref{fig:intro-illustration} (left) illustrates this issue. We are not aware of works that tested LLMs on other binary relations (beyond 'is') or other properties (beyond symmetry).
\newline\newline
Motivated by this gap in the current literature, we study how well LLMs handle \textbf{binary relations}, a core concept in math that appears frequently in most problems LLMs excel at solving when provided with sufficient in-context information \cite{ahn2024large,li2024llms,hu2025rmath}.

Our research aims to ground the extent to which LLMs can reason out-of-context, specifically focusing on logical, relational reasoning. As sketched in Figure~\ref{fig:intro-illustration}, we propose an \textbf{out-of-context representation learning} technique that introduces new tokens into a model's vocabulary and trains only their representations while leaving the other parameters unchanged.
By training only the representation of unseen tokens, out-of-context representation learning allows us to (1) understand what reasoning capabilities are present in a model and (2) without relying on external guidance, e.g., in-context learning and/or illustrations~\cite{wei2022chain,kojima2023largelanguagemodelszeroshot}. It also makes analysing the learned parameters much easier.
For the rest of the paper, we will refer to our technique as out-of-context representation learning, while (whole model) fine-tuning, which trains the entire model and is also called out-of-context learning, will be referred to as fine-tuning, to avoid confusion.
Our experiments assess the models capabilities on binary relations and their basic properties, such as reflexivity, symmetry, and transitivity.
\newline \newline
In summary, in this paper, we:
\begin{itemize}
    \item Assess the LLMs' capabilities to reason on binary relations by inferring missing pairs.
    \item Show that our technique is a more principled approach than in-context learning and fine-tuning, as it does not modify the model's parameters or provide additional information in the input prompt.
    \item Analyse the learned representations, showing that LLMs can encode useful information, such as encoding the embeddings of order relation arranged on an axis, similar to numbers.
\end{itemize}

\noindent The following sections review the current literature and formally introduce the binary relations and properties that we test.

\section{Related Work}
Several papers have explored out-of-context learning in LLMs. For example, \cite{allen2023physics} trained GPT-2 in synthetic biographies and then tested its ability to answer fine-tuning questions about specified details. The model performed well after fine-tuning on such questions for biographies not included in the test set. \cite{allen2023physics2} advanced this approach by testing questions requiring reasoning, such as determining if someone was born in an even year. While the model performed well on simple tasks (e.g., even or odd birth years) after some fine-tuning, it struggled with more complex questions requiring operations like comparison or subtraction, performing only marginally better than random guessing regardless of fine-tuning. \cite{hu2024limited,zhu2024beyond} tested similar capabilities and reported poorer results, potentially due to a lack of fine-tuning or paraphrasing in the training data. \cite{treutlein2024connecting} investigated whether LLMs could make inferences from information spread across distinct training data, concluding that LLMs can sometimes actually perform \textit{better} when fine-tuned than with in-context reasoning. \citet{berglund2023reversal} explored LLMs' difficulty with reversal relations, summarised by its title ``LLMs trained on A is B fail to learn B is A''. Recent works explain this phenomenon as an intrinsic limitation of Transformers architectures at maintaining consistent relation between the subject and the predicted object~\cite{wang2025reversalcursebindingproblem}. Similar findings are reported in the previously mentioned paper~\cite{allen2023physics2}.
Somewhat differently, but taking a more formal approach, \cite{mruthyunjaya2023rethinking} evaluates the capability of LLMs to replicate well-defined properties such as symmetry on relevant data (e.g., if a model knows that Barack Obama is married to Michelle Obama, does it know that Michelle Obama is married to Barack Obama?). However, they do not train the model on new, synthetic data, and it may well be that both directions exist in the training data. More papers took similar approaches, mainly with multi-hop reasoning \cite{yang2024large,yanaka2021exploring,welbl2018constructing,yang2018hotpotqa}.

\section{Methodology}
\label{section:methodology}
This section introduces the basic notation to describe a binary relation and an LLM. We then describe out-of-context representation training methodology and how it differs from standard fine-tuning and in-context learning. We conclude the section with a brief overview of the dataset format.

\paragraph{Binary relation.} We focus on three binary relations, namely \textbf{equality} ($=$), \textbf{inequality} ($<$), and \textbf{inclusion} ($\subset$). Each binary relation satisfies/violates several properties that are the object of our study, for example, \textbf{reflexivity}, \textbf{symmetry}, and \textbf{transitivity}, as well as other properties such as \textbf{irreflexivity}. For a finite set of elements $E$, the Cartesian product $E \bigtimes E$ identifies ordered pairs that satisfy a particular relation $\textbf{R}$.\footnote{These relations are often called \emph{homogeneous}.}

Consider, for example, equality and the set of natural numbers $\mathbb{N}$.
For any $e_1, e_2, e_3 \in \mathbb{N}$, $e_1 = e_2$ implies that $e_2 = e_1$ (symmetry); it also holds that $e_1 = e_1$ (reflexivity); finally $e_1 = e_2 \wedge e_2 = e_3 \implies e_1 = e_3$ (transitivity). 
On the other hand, $<$ preserves transitivity but fulfills irreflexivity and asymmetry, with the relation $e_1 < e_2 \wedge e_2 < e_3 \implies e_1 < e_3$ that accounts for transitivity, $e_1 < e_2 \implies e_2 \cnot< e_1$ for asymmetry, and $e_1 \cnot< e_1$ for irreflexivity.

\paragraph{Large Language Models.} An LLM is a parametrised model $\psi^{\theta}$, that maps a sequence of elements/tokens from a discrete set, namely its vocabulary $\Sigma$, into a probability distribution over the same set, i.e., $\psi^{\theta}: \Sigma^* \xrightarrow{} \mathbb{P}(\Sigma)$. The newly generated token can be appended to the input to generate longer sequences. With a small abuse of notation, we denote with $\mathbf{x}$ and $\mathbf{y}$ an input/output sequence.
We also denote with $f: \Sigma \hookrightarrow \mathbb{R}^d$ the embedding that maps each discrete token in $\Sigma$ to a real-value vector of dimension $d$.
Our settings incorporate a set of axioms $H \subset (E \bigtimes E)$ sufficient to generalise on unseen test cases $R \subset (E \bigtimes E)$, i.e., $H \models \mathbf{R}$, and a set of question in the form $e_i \mathbf{R} e_j$ the model is expected to reply for each consistently with the ground truth label $\mathbf{y}$, i.e., true or false.

\paragraph{Out-of-context representation learning.}
We augment the model’s vocabulary $\Sigma$ with new tokens—unseen during pre-training—to explicitly represent out-of-context elements, i.e., the elements of the set $E$ on which the relation $\mathbf{R}$ is defined.
Formally, for an input $\mathbf{x}$ that expresses a relation between two elements, $e_1 \mathbf{R} e_2$, and its ground truth value $\mathbf{y}$ (true/false) we aim to find:

\begin{equation}\label{eq:out-of-context}
\begin{aligned}
    & \{(e_1, \varepsilon^*_1), (e_2, \varepsilon^*_2)\} \in \argmin_{\{\varepsilon_1, \varepsilon_2\}} \mathcal{L}(\psi^{\theta}(\mathbf{x}), \mathbf{y}) 
    & \\
    & \text{s.t.} \ \ \ e_i \notin \Sigma 
    & \\
    & f(e_i) = \varepsilon_i^* \in \mathbb{R}^d 
    & \\
    & i \in \{1, 2\} \\
\end{aligned}
\end{equation}

Where $\{(e_1, \varepsilon^*_1), (e_2, \varepsilon^*_2)\}$ is the set of tokens and representations added to the model to represent the elements of the relationship in $(\mathbf{x}, \mathbf{y})$, while $\mathcal{L}$ is the model's training loss. This approach extends to multiple ordered pairs that define $H$. In practice, each token embedding is randomly initialised with its norm matching that of the other existing tokens and then optimised via gradient descent to minimise the above-reported problem \footnote{The technical details are reported in \Cref{section:technical_details}.}.

\paragraph{In-context learning.}
We represent elements in the in-context experiments using Latin alphabet letters, ensuring that each variable consists of a single existing token: no new tokens are introduced in $\Sigma$\footnote{While one can use a combination of tokens to define each variable in a binary relationship, this would introduce unnecessary complexity in tokenization and could lead to performance drops.}. The choice of the Latin vocabulary is purely conventional, and nothing prevents the use of other character systems.
For a question that tests a model's capability to infer the relation between two variables, $aRb$, we prepend the list of axioms $H$. 

A comparison of the salient characteristics of out-of-context representation learning, in-context learning, and fine-tuning is reported in Table~\ref{tab:techniques}.
  
\begin{table}
    \resizebox{.5\textwidth}{!}{%
    \begin{tabular}{c|c|c|c}
         & \small{\textbf{\makecell{In-context}}} & \small{\textbf{\makecell{Out-of-context \\ Representation}}} & \small{\textbf{Fine-tuning}}\\
         \small{\textbf{\makecell{Trained \\ Parameters}}} & $0$ & $nd$ & $|\theta|$ \\ \hline
        \small{\textbf{\makecell{Training \\ Information}}} & - & H & H \\ \hline
         
         \small{\textbf{\makecell{In-context \\ Information}}} & $\{H, \ e_i \mathbf{R} e_j\}$ & $e_i \mathbf{R} e_j$ & $e_i \mathbf{R} e_j$\\
    \end{tabular}
    }
    \caption{A comparison of the number of training parameters and amount of extra information provided in the prompt for out-of-context representation learning in-context learning, and fine-tuning. In the out-of-context setting, $n=|E|$ is the number of elements on which the relation is defined, and $d$ is the embedding dimension of the model. $H$ is an encoding of the hypotheses to correctly solve a task, while $\theta$ are the parameters of a model. $e_i \mathbf{R} e_j$ is the property the model is asked to handle properly.}
    \label{tab:techniques}
\end{table}

\paragraph{Dataset format.}

The data is formatted as a Q\&A dataset as follows:

\begin{mdframed}[]
\small{\texttt{User: Is \token{a} $\mathbf{R}$ \token{b}?}} \\
\small{\texttt{System: [Yes/No]}}
\end{mdframed}

Out-of-context representation training focuses on the final token: the model is trained to output \texttt{[Yes/No]} given the question.

As previous research suggests~\cite{allen2023physics,allen2023physics2}, we increase the question variety with paraphrases for training/test and each LLM's system prompt \footnote{\Cref{section:paraphrasing} in the appendix.}. 
In addition, to have a balanced dataset, we introduce both positive and negative questions, such as:
\begin{mdframed}[]
\small{\texttt{User: Is it wrong that \token{a} $\mathbf{R}$ \token{b}?}} \\
\small{\texttt{System: [No/Yes]}}
\end{mdframed}

In the next section, we introduce the experimental setup and the results we obtain by comparing out-of-context representation learning with in-context learning.

\section{Experiments}

In our experiments, we train Llama-3-8B, Llama-3.2-1B~\cite{grattafiori2024llama3herdmodels} and Mistral-7B-v0.3~\cite{jiang2023mistral7b} with out-of-context representation learning, as introduced in Eq.~\ref{eq:out-of-context}. The experiments for in-context learning are similar, except that the same axioms are introduced in the prompt instead of the training set, and only the minimal setting is used.

For each relation, namely strict total order, equality, and proper subset, we craft a training dataset that tests the model's capability to handle one or more properties of such a relation. The LLM is then tested on some questions that do not appear in the training, but for which the training set provides sufficient knowledge to solve them correctly.
Each evaluation contains both true and false statements (i.e., the expected ground truth answer is \texttt{[Yes/No]}), expressed with different phrasing to enhance variety.
We run each experiment 10 times with different initial random embeddings, then average the results. 
The out-of-context representation learning paradigm is implemented by introducing a new set of tokens (each paired with a dense representation), in the LLM's vocabulary, and by training only these representations.
While other works employ Chain of Thoughts~\cite{wei2022chain} to test the reasoning capabilities of LLMs~\cite{berglund2023reversal,mccoy2023embersautoregressionunderstandinglarge}, we do not employ it as the training does not contain any reasoning chain. Future works can address this limitation and check whether a model can generate chains of thought while not being explicitly trained to do so.
In the next sections, we first introduce the salient details of each binary relation alongside the implementation details; we then discuss the results of our evaluation.

\subsection{Inequality: Strict Total Order}

We test the properties of inequality with the ``smaller than'' ($<$) relation. 
We build different training sets to test whether a model can generalise on the irreflexivity ($e_1 \cnot< e_1$), asymmetry ($e_1 < e_2 \implies e_2 \cnot< e_1$), and transitivity ($e_1 < e_2 \wedge e_2 < e_3 \implies e_1 < e_3$) properties of this relation.

\paragraph{Setting I. Minimal sufficient hypotheses.}
In this setting, the model is given the minimum information required to logically derive all the answers for the test set.
The training data is the same for testing reflexivity, symmetry, and transitivity, and in the form $e_i < e_{i + 1} : 1 \le i < n$.
For irreflexivity, we test a model with pairs in the form $e_i \cnot< e_i$; pairs are in the form $e_{i + 1} \cnot< e_i$ for testing asymmetry; finally, tests are expressed in the form $e_i < e_j : j - i \ge 2$ for transitivity, which enables seeing whether the performance of the model is affected by the distance $j-i$ between the variables.

\paragraph{Setting II. Illustrative information.} 
The second setting introduces more information than is strictly necessary to derive the correct answer for test pairs. Here, when testing a certain property, the model is given in the training data, beyond the minimum information, the other properties.

For example, the transitivity training set further includes the asymmetry, i.e., $e_{i+1} \cnot< e_i : 1 \le i < n$, and the irreflexivity pairs, $e_i \cnot< e_i$. 
When testing transitivity with asymmetry given, we can test both inequality directions: $e_i < e_j $ and $e_j \cnot< e_i$ for $j - i \ge 2$. Thus, the illustrative settings allow the balancing of the number of positive and negative samples, beyond the simple "is it wrong" variation. These contrastive examples that are expected to benefit the generalisation capabilities of a model.

\subsection{Equality}
We test the equality relation by employing the ``equal to'' ($=$) relation. 
The training data is in the form $d_1 = d_2, d_2 = d_3, ..., d_{n-1} = d_n, d_n \neq e_1, e_2 = e_3, ..., e_{n-1} = e_n$ where $d_i, e_i: 1 \le i \le n$ are unique tokens introduced in the out-of-context learning procedure as per Eq.~\ref{eq:out-of-context}.
\newline 
Similarly to the strict total order, we introduce two settings: one minimal, with the training samples that specify the minimum necessary information to handle the test cases properly, and one where the training samples introduce, in addition, all properties other than the one tested.

\subsection{Inclusion: Proper Subset}
We test the properties of inclusion with the ``proper subset'' ($\subset$) relation. The training data is in the form $d_1 \subset d_2, d_2 \subset d_3, ..., d_{n-1} \subset d_n$, similarly $e_1 \subset e_2, e_2 \subset e_3, ..., e_{n-1} \subset u_n$, and finally $d_1 \cnot\subset e_n$.
\newline 
Similarly to the Strict total order and the Equality, we introduce two settings: one minimal, with the training samples that specify the minimum sufficient information to handle the test cases properly, and one where the training samples introduce properties other than the one tested.

\section{Results}

The average results over all experiments are summarised in \cref{table:average_scores}. We chose two concurrent baselines to conclude on a model's capability to handle binary relations: the accuracy of a random classifier and that of a model that predicts an input being positive or negative with the same probability as the training data distribution, regardless of the actual elements in question.
If an LLM is significantly better than both, we say the model succeeds in the task. If the model is significantly worse than both baselines, we conclude that the model has failed. Otherwise, we say that the results are inconclusive.

Hence, for the minimum variation, the overall models' performance lies between the baseline and random guess, hence are inconclusive; and the overall results for the illustrative settings are better than both the baseline and random guess, for all models. For the illustrative settings, Llama-3-8B's accuracy is better than that of Llama-3.2-1B, which is better than that of Mistral-7B-v0.3. For the minimum settings, Llama-3-8B and Mistral-7B-v0.3 score the same, and Llama-3.2-1B yields slightly worse results. A more detailed analysis of the performance on every experiment follows.

\begin{table}
\centering
\small
\begin{tabular}{@{\extracolsep{4pt}}|c|c|c|c|}
\hline
\textbf{Model} & \textbf{Settings} & \textbf{\makecell{Average \\ Accuracy}} & \textbf{Baseline} \\ \hline
\Xhline{2\arrayrulewidth}

Llama-3-8B & Minimum & 0.45 & 0.39 \\ \hline
Llama-3-8B & Illustrative & 0.65 & 0.49 \\ \hline
\Xhline{2\arrayrulewidth}
Llama-3.2-1B & Minimum & 0.43 & 0.39 \\ \hline
Llama-3.2-1B & Illustrative & 0.62 & 0.49 \\ \hline
\Xhline{2\arrayrulewidth}
Mistral-7B-v0.3 & Minimum & 0.45 & 0.39 \\ \hline
Mistral-7B-v0.3 & Illustrative & 0.6 & 0.49 \\ \hline

\end{tabular}
\caption{Average accuracy over all experiments.}
\label{table:average_scores}
\end{table}

\paragraph{Minimum information} First, in the ``minimal sufficient hypotheses'' setting, the training and test data are unbalanced by construction. For example, when testing transitivity, the training data only contains positive instances, and so is the test data. 
A success may also be caused by the model collapsing to give the same answer, no matter what the input.
We report the results in \Cref{table:llama_out_of_context_minimum,table:mistral_out_of_context_minimum,table:llama_1b_out_of_context_minimum}\footnote{Omitted tables can be found in the appendix.}. Summarising all experiments, however, all models fail in the ``minimal sufficient hypotheses'' setting and mostly follow the baseline distribution, suggesting that their best approximation of the properties comes directly from training statistics.
In other words, LLMs still struggle to generalise on well-known mathematical properties without diverse data and contrastive examples.

\paragraph{Illustrative information} The ``illustrative information'' setting mitigates the balancing issue by adding additional information that is not directly useful for solving the test cases but adds diversity and balances the datasets. The results for this setting are reported in \Cref{table:llama_out_of_context,table:mistral_out_of_context,table:llama_1b_out_of_context}: ``V'' marks a success (i.e., the model successfully learned the task and beats both the baselines), ``X'' a failure (the model performs significantly worse than both baselines), while ``?'' denotes that results are not statistically significant to conclude anything or lie in between both baselines (we conducted a T-test for comparison and a Page's trend test for trend analysis, with $\alpha=0.01$). In this setting, Llama-3-8B succeeds on all properties for all relations, except for reflexivity in the proper subset.
While slightly worse, the performance of the other models follows a similar trend.

\paragraph{In-context} When testing in-context learning, there is no training distribution because we do not train the model, so we only compare the model to a random guess. The results are reported in \Cref{table:llama_in_context,table:mistral_in_context,table:llama_1b_in_context}. For most test cases, all models perform better in out-of-context representation learning than in in-context learning, a sign that our technique improves the model capabilities while being less intrusive and more efficient than LORA~\cite{hu2021loralowrankadaptationlarge}.
\begin{figure}
    \centering
    \includegraphics[width=0.5\textwidth]{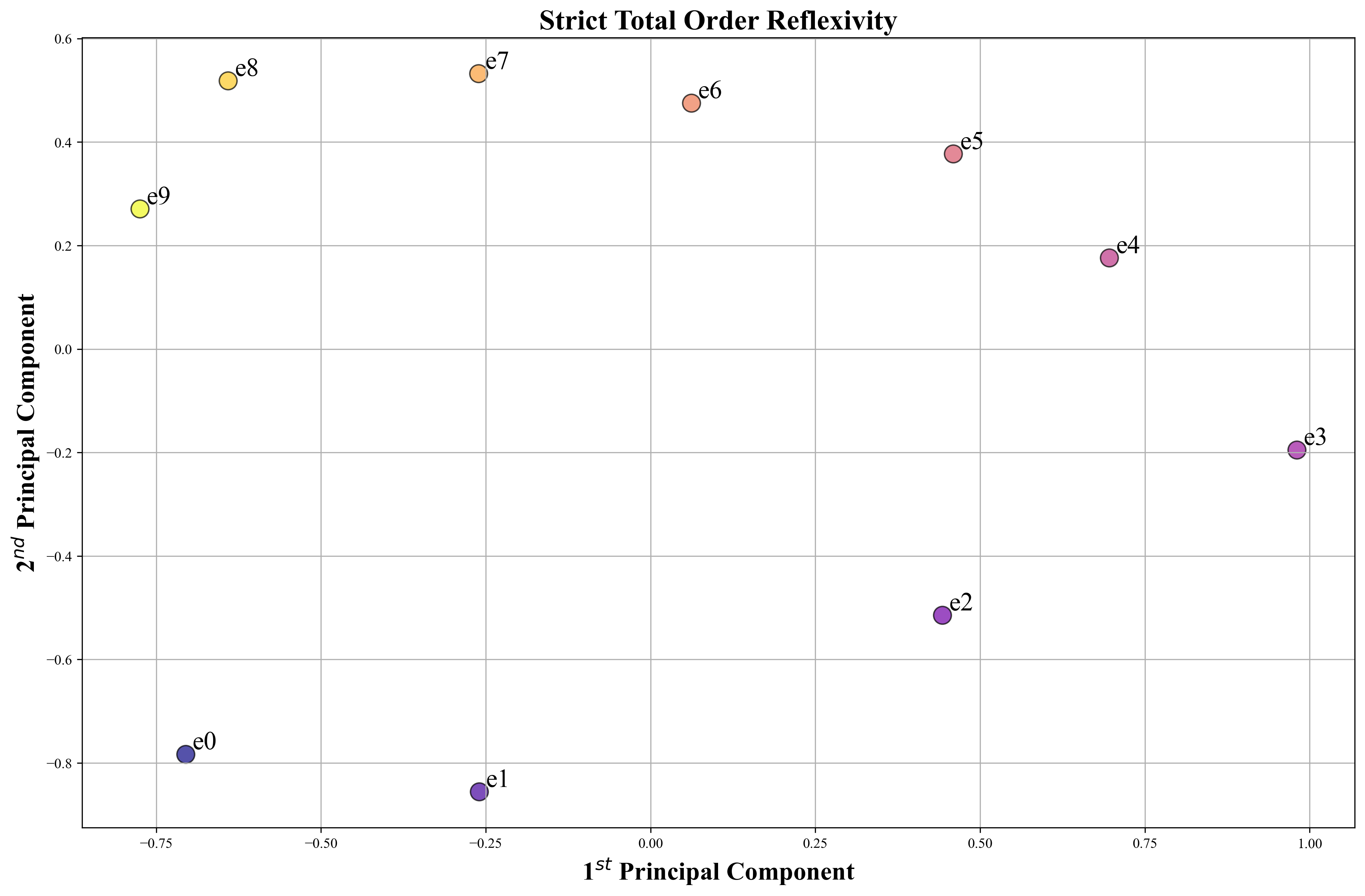}
    \caption{Llama-3-8B total order, where asymmetry and transitivity are given. The pattern where numbers appear along a circle by their order typically happens when projecting (regular) numbers embedded in Llama. The same trend is observed with the other models (\Cref{fig:mistral_total_order,fig:llama_1b_total_order}).}
    \label{fig:llama_3_total_order}
\end{figure}

\begin{figure}
    \centering
    \includegraphics[width=0.5\textwidth]{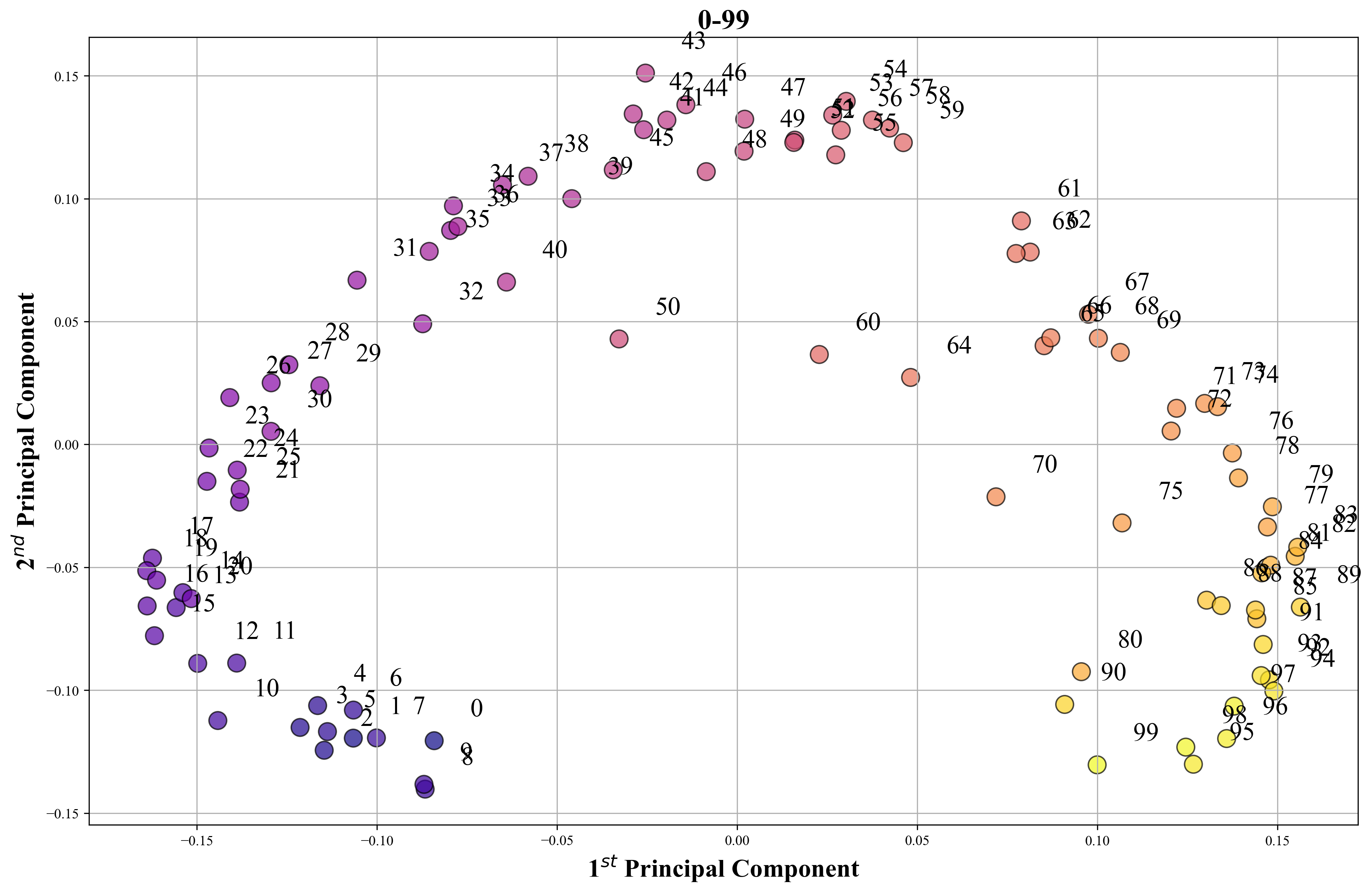}
    \caption{PCA of the embeddings of the first 100 numbers (from 0 to 99) of Llama-3-8B.}\label{fig:numbers-pca}
\end{figure}



\paragraph{The distance effect} In the total order relation, the accuracy of all tested models increases as a function of the distance between the compared symbols, the so-called \textbf{distance effect}. 
Our results confirm what was observed in~\cite{shaki2023cognitive}, though they use pre-trained tokens representing actual numbers. This effect mirrors a well-known phenomenon observed in people, who are known to respond faster and more accurately when comparing increasingly distant numbers \cite{moyer1967time,van2008dissecting,van2011origins}. 
This result is astounding in our context since the alleged number of reasoning steps needed to determine the correct answer increases as a function of the distance.
A possible explanation, which we expand on in the next paragraph, is that the models encode a fuzzy routine to compare numbers where noise plays an increasingly marginal role for distant numbers.

\paragraph{The reversal curse.} 
Another interesting phenomenon that our experiments explain is that of the \textbf{reversal curse}, i.e., a model that fails to generalise ``B is A'' after learning ``A is B''~\cite{berglund2023reversal}. 
Our experiments show that Llama-3-8B and Mistral-7B-v0.3 (small models compared to larger LLMs such as LLama-3.1-405B) successfully learn symmetry (both in the minimum settings, and Llama-3-8B also in the illustrative settings).
We argue that the reversal curse arises from the linguistic ambiguity of 'is', which can signal equality or function as a copula in noun phrases. With proper training, as in our out-of-context representation learning, small models succeed at the task and are unaffected by this issue.

When tested with in-context learning, i.e., the training data is instead provided as part of the prompt, Llama-3-8B succeeds mostly on equality. Surprisingly, Mistral-7B-v0.3 succeeds mostly on the strict total order and proper subset, except on transitivity, where the model fails. Llama-3.2-1B achieves low accuracy, even when performing statistically significantly better than random guess, stressing the role of model size in this setting.

\paragraph{Learned representations.}
We analyse the learned representations for each experiment in the ``illustrative information'' setting, i.e., when the models mostly succeed in the task, with a one or two-dimensional PCA to reveal the dimensions of the maximum variation.
As reported in Figure~\ref{fig:llama_3_total_order}, the projection of the newly introduced representations resembles that of the first $100$ natural numbers, as per Figure~\ref{fig:numbers-pca}.

\begin{figure}
    \centering
    \includegraphics[width=0.5\textwidth]{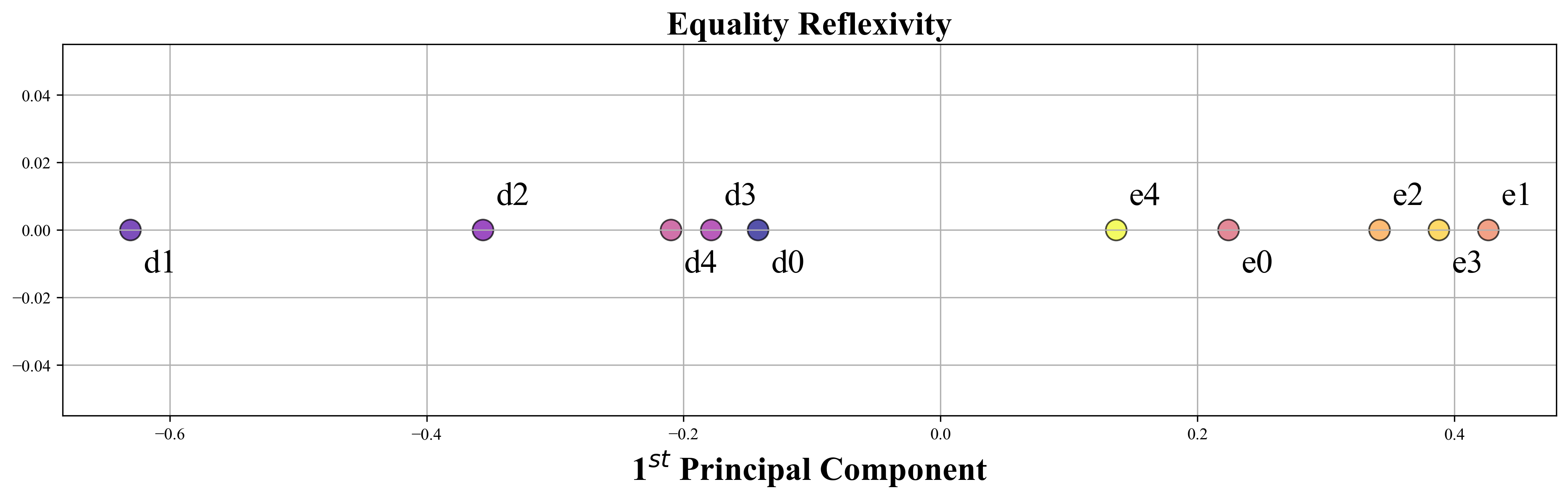}
    \caption{Llama-3-8B, equality, where symmetry and transitivity are given. The equivalence classes are clear. This also happens when reflexivity and transitivity are given. The same trend is observed with the other models (\Cref{fig:mistral_equality,fig:llama_1b_equality}).}\label{fig:pca-equality}
\end{figure}

\begin{figure}
    \centering
    \includegraphics[width=0.5\textwidth]{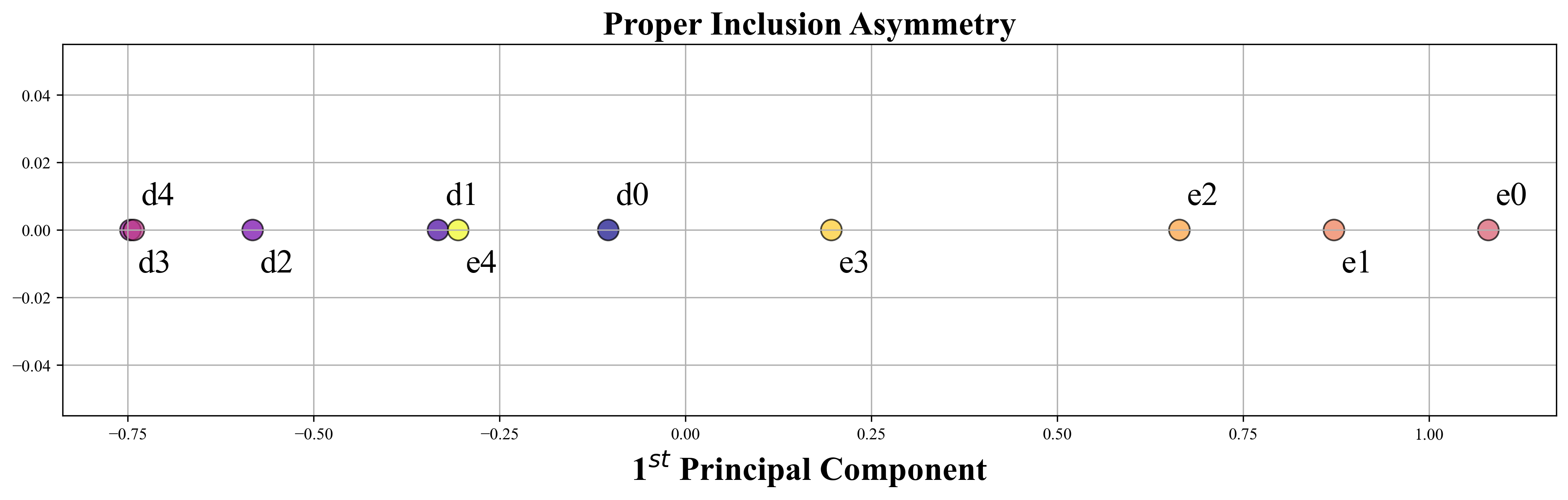}
    \caption{Llama-3-8B proper subset, where irreflexivity and transitivity are given. Groups that are contained by others are to their right. v0 is also to the right of v9, even though it is explicitly stated in the training data that it is not strictly contained in v9. A similar trend is observed with the other models (\Cref{fig:mistral_subset,fig:llama_1b_subset}).}\label{fig:pca-proper-subset}
\end{figure}

We hypothesise that for the total order relation, Llama learns to model asymmetry and transitivity similarly to how it does for natural numbers (i.e., by projecting the embeddings into a low-dimensional manifold that satisfies the two properties). We also observe similar representations in Mistral's embeddings.
On the other hand, symmetry and transitivity of the equality relation, as well as irreflexivity and transitivity of the proper subset relation, require a more straightforward representation, as per Figures~\ref{fig:pca-equality} and~\ref{fig:pca-proper-subset}. In this sense, out-of-context representation learning is efficient and suggests the existence of shared learning dynamics for similar problems/representations.

The graph analysis is that of the averaged learned embeddings over the multiple iterations we ran for each experiment. The patterns are not observed directly for a single iteration. 
We also note that the accuracy of using these average representations (\Cref{table:llama_out_of_context_average,table:mistral_out_of_context_average,table:llama_1b_out_of_context_average}) are similar to the average on each learned representation (average difference of +0.01 for Llama-3-8B, -0.02 for Mistral-7B-v0.3, and -0.06 for Llama-3.2-1B). This is a known phenomena that occur when averaging models' weights \cite{rame2022diverse,izmailov2018averaging,cha2021swad}, especially when the test data is out-of-distribution \cite{rame2022diverse}, as in our case.

\section*{Limitations and Open Questions}
While in-context learning provides relevant information in the input prompt, fine-tuning modifies the weights of a model.
The former tests the capability of a model to reason with external information; the latter optimises the model's parameters and tests whether a model can learn such a property; yet, both in-context learning and fine-tuning are prone to the bias of pre-existing tokens (see the discussion on the ``reversal curse''), and fine-tuning can also incur overfitting.
On the other hand, out-of-context representation learning does not provide external information or change the model's parameters. As long as one can introduce new tokens in a model, our technique serves as a way to assess a model's capability on a task.

While in many cases our approach succeeds and supports the hypothesis that LLMs can properly reason about binary relations, it raises some questions when they fail.
In particular, Tables~\ref{table:llama_out_of_context} and~\ref{table:mistral_out_of_context} (marked with a ``?'' symbol) show that models behave ambiguously for asymmetry in strict total order, unless the elements involved are the farther as per the initial hypotheses.  
Results support the hypothesis that the embedding representations learnt with our technique are noisy (see Figures~\ref{fig:pca-equality} and~\ref{fig:pca-proper-subset}) and thus subject to errors for short-distance comparisons.

\section{Conclusions}
This paper explores the ability of LLMs to reason about binary relationships through out-of-context representation learning. 
We assessed whether LLMs can generalise reasoning beyond in-context learning by examining relational properties such as reflexivity, symmetry, and transitivity, on knowledge the model encountered only during training. 
Our findings indicate that out-of-context representation learning allows for better generalisation in most tasks we tested. We show that when the models succeed, they do so by arranging the learned embeddings according to the task.

Future research will test the robustness of out-of-context representation learning against data contamination by repeating the experiments on model trained on plain text version of our experiments.

\clearpage

\begin{table*}[h]
\centering
\small
\begin{tabular}{@{\extracolsep{4pt}}|c|c|c|c|}
\hline
\textbf{Relation} & \textbf{Property} & \textbf{Accuracy} & \textbf{Baseline} \\ \hline

\Xhline{2\arrayrulewidth}

\textbf{\makecell{Strict \\ Total Order}} & Irreflexivity & 0.09 & 0 \\  \hline
\textbf{\makecell{Strict \\ Total Order}} & \makecell{Asymmetry} & 0.01 & 0 \\ \hline
\textbf{\makecell{Strict \\ Total Order}} & \makecell{Transitivity \\ \{2, 3, 4, 5, 6, 7, 8, 9\} hops} & 0.97, 0.98, 0.99, 0.97, 0.99, 0.97, 0.99, 0.97 & 1 \\ \hline

\Xhline{2\arrayrulewidth}

\textbf{Equality} & Reflexivity & 0.86 & 0.89 \\  \hline
\textbf{Equality} & Symmetry & 0.75 & 0.5 \\  \hline
\textbf{Equality} & \makecell{Transitivity \\ \{2, 3, 4\} hops} & 0.62, 0.6, 0.46 & 0.5 \\ \hline

\Xhline{2\arrayrulewidth}

\textbf{Proper Subset} & Irreflexivity & 0.2 & 0.05 \\ \hline
\textbf{Proper Subset} & \makecell{Asymmetry} & 0.07 & 0.05 \\ \hline
\textbf{Proper Subset} & \makecell{Transitivity \\ \{2, 3, 4\} hops} & 0.59, 0.52, 0.42 & 0.5 \\ \hline

\end{tabular}
\caption{Results for Llama-3-8B, out-of-context representation learning, minimum information settings.}
\label{table:llama_out_of_context_minimum}
\end{table*}

\begin{table*}[h]
\centering
\small
\begin{tabular}{@{\extracolsep{4pt}}|c|c|c|c|c|}
\hline
\textbf{Relation} & \textbf{Property} & \textbf{Accuracy} & \textbf{Success} & \textbf{Trend} \\ \hline

\Xhline{2\arrayrulewidth}

\textbf{\makecell{Strict \\ Total Order}} & Irreflexivity & 0.58 & V & None \\ \hline
\textbf{\makecell{Strict \\ Total Order}} & \makecell{Asymmetry \\ distance of \{1, 2, 3, 4, 5, 6, 7, 8, 9\}} & \makecell{0.12, 0.17, 0.19, 0.32, \\ 0.4, 0.45, 0.56, 0.78, 0.97} & \makecell{?, ?, ?, ?, \\ ?, ?, ?, V, V} & Increasing \\ \hline
\textbf{\makecell{Strict \\ Total Order}} & \makecell{Transitivity \\ \{2, 3, 4, 5, 6, 7, 8, 9\} hops} & \makecell{0.61, 0.55, 0.61, 0.65, \\ 0.65, 0.76, 0.9, 0.9} & \makecell{V, ?, V, V, \\ V, V, V, V} & Increasing \\ \hline

\Xhline{2\arrayrulewidth}

\textbf{Equality} & Reflexivity & 0.98 & V & None \\ \hline
\textbf{Equality} & \makecell{Average symmetry \\ distance of \{1, 2, 3, 4\} } & 0.62, 0.63, 0.68, 0.72 & V, V, V, V & Increasing \\ \hline
\textbf{Equality} & \makecell{Average transitivity \\ \{2, 3, 4\} hops} & 0.57, 0.55, 0.45 & V, ?, ? & Decreasing \\ \hline

\Xhline{2\arrayrulewidth}

\textbf{Proper Subset} & Irreflexivity & 0.45 & ? & None \\ \hline
\textbf{Proper Subset} & \makecell{Asymmetry \\ distance of \{1, 2, 3, 4\}} & 0.65, 0.82, 0.89, 0.94 & ?, V, V, V & Increasing \\ \hline
\textbf{Proper Subset} & \makecell{Average transitivity \\ \{2, 3, 4\} hops} & 0.66, 0.63, 0.68 & V, V, V & Not found \\ \hline

\end{tabular}
\caption{Results for Llama-3-8B, out-of-context representation learning, illustrative information settings. Symbol ``V'' marks a success (i.e., the model successfully learned the task and beats both the baselines), ``X'' a failure (the model failed on both baselines), while ``?'' denotes that results are not statistically significant to conclude anything (we conducted a T-test for comparison and a Page's trend test for trend analysis).}
\label{table:llama_out_of_context}
\end{table*}

\begin{table*}[h]
\centering
\small
\begin{tabular}{@{\extracolsep{4pt}}|c|c|c|c|c|}
\hline
\textbf{Relation} & \textbf{Property} & \textbf{Accuracy} & \textbf{Success} & \textbf{Trend} \\ \hline

\Xhline{2\arrayrulewidth}

\textbf{\makecell{Strict \\ Total Order}} & Irreflexivity & 0.55 & V & None \\ \hline
\textbf{\makecell{Strict \\ Total Order}} & Asymmetry & 0.48 & X & None \\ \hline
\textbf{\makecell{Strict \\ Total Order}} & \makecell{Transitivity \\ \{2, 3, 4, 5, 6, 7, 8, 9\} hops} & \makecell{0.39, 0.37, 0.37, 0.38, \\ 0.37, 0.38, 0.38, 0.37} & \makecell{X, X, X, X, \\ X, X, X, X} & Not found \\ \hline

\Xhline{2\arrayrulewidth}

\textbf{Equality} & Reflexivity & 0.9 & V & None \\ \hline
\textbf{Equality} & Symmetry & 0.81 & V & None \\ \hline
\textbf{Equality} & \makecell{Transitivity \\ 2, 3, 4 hops} & 0.6, 0.51, 0.45 & V, ?, X & Decreasing \\ \hline

\Xhline{2\arrayrulewidth}

\textbf{Proper Subset} & Irreflexivity & 0.47 & X & None \\ \hline
\textbf{Proper Subset} & Asymmetry & 0.58 & V & None \\ \hline
\textbf{Proper Subset} & \makecell{Transitivity \\ \{2, 3, 4\} hops} & 0.5, 0.5, 0.5 & ?, ?, ? & Not found \\ \hline

\end{tabular}
\caption{Results for Llama-3-8B, in-context learning. Symbol ``V'' marks a success (i.e., the model successfully learned the task and beats random guess), ``X'' a failure, while ``?'' denotes that results are not statistically significant to conclude anything (we conducted a T-test for comparison and a Page's trend test for trend analysis).}
\label{table:llama_in_context}
\end{table*}





\clearpage

\bibliography{main}

\clearpage

\section{Appendix}

Our code and results can be found in \href{github.com/1jonathan123/out-of-context-representation-learning}{github.com/1jonathan123/out-of-context-representation-learning}.

\subsection{Training and Evaluation}
\label{section:technical_details}

For training, we used AdamW with the default learning rate in pytorch (0.001). For weight decay, since the embedding dimension is large compared to the amount of training data, we aim to use the largest weight decay possible where the model still achieves a loss of zero on the training data. Practically, this means beginning with a large weight decay (0.1) and if the model doesn't achieve a zero loss on the training data after $20$ iterations, we try again with a weight decay reduced by a half, and repeat this process until the model succeeds perfectly on the training data.

We performed the training on a single H-200 GPU, which takes about 10 days for running all experiments on all models.

For evaluation, the accuracy of the model is defined as the probability of predicting the correct answer (Yes/No), divided by the total probability of predicting either "Yes" or "No". Since we examine the output distribution directly, no sampling algorithm is used.

\subsection{Paraphrasing}
\label{section:paraphrasing}
As reported in \Cref{section:methodology}, we use multiple paraphrasing for both in-context and out-of-context learning to increase the variety of the training data.

For the system prompt, a text that instructs the LLM how to behave and typically appears as the first message, the possible variations are:

\begin{mdframed}[]
\small{\texttt{You are a helpful and honest assistant. Please, respond concisely and truthfully. Limit your answers to a single word when possible.}}
\end{mdframed}

\begin{mdframed}[]
\small{\texttt{You are a reliable and straightforward assistant. Respond briefly and accurately. Aim for single-word answers when feasible.}}
\end{mdframed}

\begin{mdframed}[]
\small{\texttt{You are a dependable and truthful assistant. Please reply succinctly and honestly, using one word whenever possible.}}
\end{mdframed}

\begin{mdframed}[]
\small{\texttt{You are a trustworthy and concise assistant. Keep your answers brief and truthful, using a single word if you can.}}
\end{mdframed}

\begin{mdframed}[]
\small{\texttt{You are an efficient and honest assistant. Provide concise, truthful responses, limiting them to one word when appropriate.}}
\end{mdframed}

\begin{mdframed}[]
\small{\texttt{You are a helpful and sincere assistant. Reply concisely and truthfully. Use a single word whenever possible.}}
\end{mdframed}

\begin{mdframed}[]
\small{\texttt{You are a clear and honest assistant. Please respond with brevity and accuracy, sticking to one word if it fits.}}
\end{mdframed}

And for each setting and two variables $a, b$ being tested/trained on:

\subsubsection{Strict Total Order}

The positive variations are:

\begin{mdframed}[]
\small{\texttt{Is \token{a} [comparison] \token{b}?}}
\end{mdframed}

\begin{mdframed}[]
\small{\texttt{Is it true that \token{a} [comparison] \token{b}?}}
\end{mdframed}

And the negative:

\begin{mdframed}[]
\small{\texttt{Is it wrong that \token{a} [comparison] \token{b}?}}
\end{mdframed}

\begin{mdframed}[]
\small{\texttt{Is it false that \token{a} [comparison] \token{b}?}}
\end{mdframed}

Where [comparison] is either: "<", "is smaller than", "is less than".

\subsubsection{Equality}

The positive variations are:

\begin{mdframed}[]
\small{\texttt{Is \token{a} equal to \token{b}?}}
\end{mdframed}

\begin{mdframed}[]
\small{\texttt{Does \token{a} equal \token{b}?}}
\end{mdframed}

\begin{mdframed}[]
\small{\texttt{Does \token{a} = \token{b}?}}
\end{mdframed}

\begin{mdframed}[]
\small{\texttt{Does \token{a} have the same value as \token{b}?}}
\end{mdframed}

And the negative:

\begin{mdframed}[]
\small{\texttt{Is it wrong that \token{a} is equal to \token{b}?}}
\end{mdframed}

\begin{mdframed}[]
\small{\texttt{Is it wrong that \token{a} equals \token{b}?}}
\end{mdframed}

\begin{mdframed}[]
\small{\texttt{Is it wrong that \token{a} = \token{b}?}}
\end{mdframed}

\begin{mdframed}[]
\small{\texttt{Is it wrong that \token{a} has the same value as \token{b}?}}
\end{mdframed}

\subsubsection{Proper Subset}

\begin{mdframed}[]
\small{\texttt{Is \token{a} a strict subset of \token{b}?}}
\end{mdframed}

\begin{mdframed}[]
\small{\texttt{Is \token{a} strictly a subset of \token{b}?}}
\end{mdframed}

\begin{mdframed}[]
\small{\texttt{Is \token{a} strictly contained in \token{b}?}}
\end{mdframed}

\begin{mdframed}[]
\small{\texttt{Is \token{a} a proper subset of \token{b}?}}
\end{mdframed}

And the negative:

\begin{mdframed}[]
\small{\texttt{Is it wrong that \token{a} is a strict subset of \token{b}?}}
\end{mdframed}

\begin{mdframed}[]
\small{\texttt{Is it wrong that \token{a} is strictly a subset of \token{b}?}}
\end{mdframed}

\begin{mdframed}[]
\small{\texttt{Is it wrong that \token{a} is strictly contained in \token{b}?}}
\end{mdframed}

\begin{mdframed}[]
\small{\texttt{Is it wrong that \token{a} is a proper subset of \token{b}?}}
\end{mdframed}

\subsection{Mistral and Llama-1b Results}

The results for mistral and llama-1b, for all training settings, are reported in \Cref{table:mistral_in_context,table:mistral_out_of_context,table:mistral_out_of_context_minimum,table:llama_1b_in_context,table:llama_1b_out_of_context,table:llama_1b_out_of_context_minimum}.

\begin{table*}[h]
\centering
\small
\begin{tabular}{@{\extracolsep{4pt}}|c|c|c|c|}
\hline
\textbf{Relation} & \textbf{Property} & \textbf{Accuracy} & \textbf{Baseline} \\ \hline

\Xhline{2\arrayrulewidth}

\textbf{\makecell{Strict \\ Total Order}} & Irreflexivity & 0.17 & 0 \\  \hline
\textbf{\makecell{Strict \\ Total Order}} & \makecell{Asymmetry} & 0.03 & 0 \\ \hline
\textbf{\makecell{Strict \\ Total Order}} & \makecell{Transitivity \\ \{2, 3, 4, 5, 6, 7, 8, 9\} hops} & 0.85, 0.97, 0.89, 0.93, 0.91, 0.92, 0.96, 0.84 & 1 \\ \hline

\Xhline{2\arrayrulewidth}

\textbf{Equality} & Reflexivity & 0.82 & 0.89 \\  \hline
\textbf{Equality} & Symmetry & 0.65 & 0.5 \\  \hline
\textbf{Equality} & \makecell{Transitivity \\ \{2, 3, 4\} hops} & 0.65, 0.63, 0.51 & 0.5 \\ \hline

\Xhline{2\arrayrulewidth}

\textbf{Proper Subset} & Irreflexivity & 0.25 & 0.05 \\ \hline
\textbf{Proper Subset} & \makecell{Asymmetry} & 0.14 & 0.05 \\ \hline
\textbf{Proper Subset} & \makecell{Transitivity \\ \{2, 3, 4\} hops} & 0.57, 0.5, 0.41 & 0.5 \\ \hline

\end{tabular}
\caption{Results for Mistral-7B-v0.3, out-of-context representation learning, minimum information settings.}\label{table:mistral_out_of_context_minimum}
\end{table*}

\begin{table*}[h]
\centering
\small
\begin{tabular}{@{\extracolsep{4pt}}|c|c|c|c|c|}
\hline
\textbf{Relation} & \textbf{Property} & \textbf{Accuracy} & \textbf{Success} & \textbf{Trend} \\ \hline

\Xhline{2\arrayrulewidth}

\textbf{\makecell{Strict \\ Total Order}} & Irreflexivity & 0.46 & ? & None \\ \hline
\textbf{\makecell{Strict \\ Total Order}} & \makecell{Asymmetry \\ distance of \{1, 2, 3, 4, 5, 6, 7, 8, 9\}} & \makecell{0.17, 0.24, 0.28, 0.35, \\ 0.39, 0.4, 0.55, 0.65, 0.77} & \makecell{?, ?, ?, ?, \\ ?, ?, ?, V, V} & Increasing \\ \hline
\textbf{\makecell{Strict \\ Total Order}} & \makecell{Transitivity \\ \{2, 3, 4, 5, 6, 7, 8, 9\} hops} & \makecell{0.56, 0.58, 0.52, 0.58, \\ 0.65, 0.64, 0.75, 0.94} & \makecell{?, V, ?, ?, \\ V, V, V, V} & Increasing \\ \hline

\Xhline{2\arrayrulewidth}

\textbf{Equality} & Reflexivity & 0.94 & V & None \\ \hline
\textbf{Equality} & \makecell{Average symmetry \\ distance of 1, 2, 3, 4} & 0.48, 0.49, 0.46, 0.4 & ?, ?, ?, X & Decreasing \\ \hline
\textbf{Equality} & \makecell{Average transitivity \\ \{2, 3, 4\} hops} & 0.52, 0.51, 0.45 & ?, V, ? & Not found \\ \hline

\Xhline{2\arrayrulewidth}

\textbf{Proper Subset} & Irreflexivity & 0.56 & ? & None \\ \hline
\textbf{Proper Subset} & \makecell{Asymmetry \\ distance of \{1, 2, 3, 4\}} & 0.73, 0.86, 0.93, 0.99 & V, V, V, V & Increasing \\ \hline
\textbf{Proper Subset} & \makecell{Average transitivity \\ \{2, 3, 4\} hops} & 0.57, 0.54, 0.62 & V, ?, V & Not found \\ \hline

\end{tabular}
\caption{Results for Mistral-7B-v0.3, out-of-context representation learning, illustrative settings. Symbol ``V'' marks a success (i.e., the model successfully learned the task and beats both the baselines), ``X'' a failure (the model failed on both baselines), while ``?'' denotes that results are not statistically significant to conclude anything (we conducted a T-test for comparison and a Page's trend test for trend analysis).}\label{table:mistral_out_of_context}
\end{table*}

\begin{table*}[h]
\centering
\small
\begin{tabular}{@{\extracolsep{4pt}}|c|c|c|c|c|}
\hline
\textbf{Relation} & \textbf{Property} & \textbf{Accuracy} & \textbf{Success} & \textbf{Trend} \\ \hline

\Xhline{2\arrayrulewidth}

\textbf{\makecell{Strict \\ Total Order}} & Irreflexivity & 0.81 & V & None \\ \hline
\textbf{\makecell{Strict \\ Total Order}} & Asymmetry & 0.55 & V & None \\ \hline
\textbf{\makecell{Strict \\ Total Order}} & \makecell{Transitivity \\ \{2, 3, 4, 5, 6, 7, 8, 9\} hops} & \makecell{0.44, 0.4, 0.41, 0.39, \\ 0.38, 0.41, 0.42, 0.38} & \makecell{X, X, X, X, \\ X, X, X, X} & Decreasing \\ \hline

\Xhline{2\arrayrulewidth}

\textbf{Equality} & Reflexivity & 0.41 & X & None \\ \hline
\textbf{Equality} & Symmetry & 0.52 & ? & None \\ \hline
\textbf{Equality} & \makecell{Transitivity \\ \{2, 3, 4\} hops} & 0.53, 0.51, 0.5 & ?, ?, ? & Not found \\ \hline

\Xhline{2\arrayrulewidth}

\textbf{Proper Subset} & Irreflexivity & 0.83 & V & None \\ \hline
\textbf{Proper Subset} & Asymmetry & 0.79 & V & None \\ \hline
\textbf{Proper Subset} & \makecell{Transitivity \\ \{2, 3, 4\} hops} & 0.5, 0.49, 0.49 & ?, ?, ? & Not found \\ \hline

\end{tabular}
\caption{Results for Mistral-7B-v0.3, in-context learning. Symbol ``V'' marks a success (i.e., the model successfully learned the task and beats random guess), ``X'' a failure, while ``?'' denotes that results are not statistically significant to conclude anything (we conducted a T-test for comparison and a Page's trend test for trend analysis).}
\label{table:mistral_in_context}
\end{table*}

\begin{table*}[h]
\centering
\small
\begin{tabular}{@{\extracolsep{4pt}}|c|c|c|c|}
\hline
\textbf{Relation} & \textbf{Property} & \textbf{Accuracy} & \textbf{Baseline} \\ \hline

\Xhline{2\arrayrulewidth}

\textbf{\makecell{Strict \\ Total Order}} & Irreflexivity & 0.15 & 0 \\  \hline
\textbf{\makecell{Strict \\ Total Order}} & \makecell{Asymmetry} & 0.02 & 0 \\ \hline
\textbf{\makecell{Strict \\ Total Order}} & \makecell{Transitivity \\ \{2, 3, 4, 5, 6, 7, 8, 9\} hops} & \makecell{0.9, 0.97, 0.93, 0.95, \\ 0.93, 0.95, 0.91, 0.92} & 1 \\ \hline

\Xhline{2\arrayrulewidth}

\textbf{Equality} & Reflexivity & 0.79 & 0.89 \\  \hline
\textbf{Equality} & Symmetry & 0.58 & 0.5 \\  \hline
\textbf{Equality} & \makecell{Average Transitivity \\ \{2, 3, 4\} hops} & 0.63, 0.57, 0.52 & 0.5 \\ \hline

\Xhline{2\arrayrulewidth}

\textbf{Proper Subset} & Irreflexivity & 0.22 & 0.05 \\ \hline
\textbf{Proper Subset} & \makecell{Asymmetry} & 0.09 & 0.05 \\ \hline
\textbf{Proper Subset} & \makecell{Average Transitivity \\ \{2, 3, 4\} hops} & 0.55, 0.54, 0.42 & 0.5 \\ \hline

\end{tabular}
\caption{Results for Llama-3.2-1B, out-of-context representation learning, minimum information settings.}
\label{table:llama_1b_out_of_context_minimum}
\end{table*}

\begin{table*}[h]
\centering
\small
\begin{tabular}{@{\extracolsep{4pt}}|c|c|c|c|c|}
\hline
\textbf{Relation} & \textbf{Property} & \textbf{Accuracy} & \textbf{Success} & \textbf{Trend} \\ \hline

\Xhline{2\arrayrulewidth}

\textbf{\makecell{Strict \\ Total Order}} & Irreflexivity & 0.55 & V & None \\ \hline
\textbf{\makecell{Strict \\ Total Order}} & \makecell{Asymmetry \\ distance of \makecell{\{1, 2, 3, 4, \\ 5, 6, 7, 8, 9\}}} & \makecell{0.16, 0.21, 0.29, 0.32, \\ 0.35, 0.45, 0.6, 0.82, 0.95} & \makecell{?, ?, ?, ?, \\ ?, ?, ?, V, V} & Increasing \\ \hline
\textbf{\makecell{Strict \\ Total Order}} & \makecell{Transitivity \\ \makecell{\{2, 3, 4, 5, \\ 6, 7, 8, 9\}} hops} & \makecell{0.56, 0.55, 0.58, 0.61, \\ 0.65, 0.69, 0.74, 0.81} & \makecell{V, ?, V, V, \\ V, V, V, V} & Increasing \\ \hline

\Xhline{2\arrayrulewidth}

\textbf{Equality} & Reflexivity & 0.91 & V & None \\ \hline
\textbf{Equality} & \makecell{Average symmetry \\ distance of \{1, 2, 3, 4\}} & 0.5, 0.48, 0.47, 0.44 & ?, ?, ?, ? & Decreasing \\ \hline
\textbf{Equality} & \makecell{Average Transitivity \\ \{2, 3, 4\} hops} & 0.56, 0.51, 0.5 & V, ?, ? & Decreasing \\ \hline

\Xhline{2\arrayrulewidth}

\textbf{Proper Subset} & Irreflexivity & 0.58& ? & None \\ \hline
\textbf{Proper Subset} & \makecell{Asymmetry \\ distance of \{1, 2, 3, 4\}} & 0.67, 0.8, 0.92, 0.97 & V, V, V, V & Increasing \\ \hline
\textbf{Proper Subset} & \makecell{Average Transitivity \\ \{2, 3, 4\} hops} & 0.59, 0.62, 0.67 & V, V, V & Not found \\ \hline

\end{tabular}
\caption{Results for Llama-3.2-1B, out-of-context representation learning, illustrative information settings. Symbol ``V'' marks a success (i.e., the model successfully learned the task and beats both the baselines), ``X'' a failure (the model failed on both baselines), while ``?'' denotes that results are not statistically significant to conclude anything (we conducted a T-test for comparison and a Page's trend test for trend analysis).}
\label{table:llama_1b_out_of_context}
\end{table*}

\begin{table*}[h]
\centering
\small
\begin{tabular}{@{\extracolsep{4pt}}|c|c|c|c|c|}
\hline
\textbf{Relation} & \textbf{Property} & \textbf{Accuracy} & \textbf{Success} & \textbf{Trend} \\ \hline

\Xhline{2\arrayrulewidth}

\textbf{\makecell{Strict \\ Total Order}} & Irreflexivity & 0.56 & V & None \\ \hline
\textbf{\makecell{Strict \\ Total Order}} & Asymmetry & 0.51 & V & None \\ \hline
\textbf{\makecell{Strict \\ Total Order}} & \makecell{Transitivity \\ \makecell{\{2, 3, 4, 5, \\ 6, 7, 8, 9\}} hops} & \makecell{0.49, 0.49, 0.49, 0.49, \\ 0.48, 0.48, 0.49, 0.49} & \makecell{X, X, X, X, \\ X, X, X, ?} & Not found \\ \hline

\Xhline{2\arrayrulewidth}

\textbf{Equality} & Reflexivity & 0.53 & V & None \\ \hline
\textbf{Equality} & Symmetry & 0.5& ? & None \\ \hline
\textbf{Equality} & \makecell{Transitivity \\ \{2, 3, 4\} hops} & 0.5, 0.5, 0.5 & ?, ?, ?& Not found \\ \hline

\Xhline{2\arrayrulewidth}

\textbf{Proper Subset} & Irreflexivity & 0.51 & V & None \\ \hline
\textbf{Proper Subset} & Asymmetry & 0.52 & V & None \\ \hline
\textbf{Proper Subset} & \makecell{Transitivity \\ \{2, 3, 4\} hops} & 0.5, 0.5, 0.5 & ?, ?, ? & Not found \\ \hline

\end{tabular}
\caption{Results for Llama-3.2-1B, in-context learning. Symbol ``V'' marks a success (i.e., the model successfully learned the task and beats random guess), ``X'' a failure, while ``?'' denotes that results are not statistically significant to conclude anything (we conducted a T-test for comparison and a Page's trend test for trend analysis).}
\label{table:llama_1b_in_context}
\end{table*}

\subsection{Averaged Tokens Results}

We report in \Cref{table:llama_1b_out_of_context_average,table:llama_out_of_context_average,table:mistral_out_of_context_average} the results for the averaged learned representations. Each out-of-context experiment was run 10 times, resulting in 10 learned representations. We average them all and then test the model in the standard manner.

\begin{table*}[h]
\centering
\small
\begin{tabular}{@{\extracolsep{4pt}}|c|c|c|c|c|}
\hline
\textbf{Relation} & \textbf{Property} & \textbf{Other properties are given} & \textbf{Accuracy} & \textbf{Baseline} \\ \hline

\Xhline{2\arrayrulewidth}

\textbf{\makecell{Strict \\ Total order}} & Irreflexivity & No & 0.12 & 0 \\  \hline
\textbf{\makecell{Strict \\ Total order}} & \makecell{Asymmetry} & No & 0.03 & 0 \\ \hline
\textbf{\makecell{Strict \\ Total order}} & \makecell{Transitivity \\ \{2, 3, 4, 5, 6, 7, 8, 9\} hops} & No & \makecell{0.99, 0.98, 0.99, 0.95, \\ 0.97, 0.98, 0.96, 0.98} & 1 \\ \hline

\textbf{\makecell{Strict \\ Total order}} & Irreflexivity & Yes & 0.65 & 0.5 \\  \hline
\textbf{\makecell{Strict \\ Total order}} & \makecell{Asymmetry \\ distance of \{1, 2, 3, 4, 5, 6, 7, 8, 9\}} & Yes & \makecell{0.18, 0.18, 0.38, 0.26, \\ 0.58, 0.59, 0.46, 0.96, 1.0} & 0.18 \\ \hline
\textbf{\makecell{Strict \\ Total order}} & \makecell{Transitivity \\ \{2, 3, 4, 5, 6, 7, 8, 9\} hops} & Yes & \makecell{0.69, 0.59, 0.62, 0.61, \\ 0.71, 0.9, 0.96, 1.0} & 0.5 \\ \hline

\Xhline{2\arrayrulewidth}

\textbf{Equality} & Reflexivity & No & 0.95 & 0.89 \\  \hline
\textbf{Equality} & Average symmetry & No & 0.47 & 0.5 \\  \hline
\textbf{Equality} & \makecell{Average transitivity \\ \{2, 3, 4\} hops} & No & 0.54, 0.63, 0.58 & 0.5 \\ \hline

\textbf{Equality} & Reflexivity & Yes & 1.0 & 0.44 \\  \hline
\textbf{Equality} & \makecell{Average symmetry \\ \{1, 2, 3, 4\} hops} & Yes & 0.79, 0.86, 0.87, 0.89 & 0.5 \\ \hline
\textbf{Equality} & \makecell{Average transitivity \\ \{2, 3, 4\} hops} & Yes & 0.51, 0.49, 0.5 & 0.5 \\ \hline

\Xhline{2\arrayrulewidth}

\textbf{Proper Subset} & Irreflexivity & No & 0.15 & 0.05 \\ \hline
\textbf{Proper Subset} & \makecell{Asymmetry} & No & 0.07 & 0.05 \\ \hline
\textbf{Proper Subset} & \makecell{Average transitivity \\ \{2, 3, 4\} hops} & No & 0.53, 0.41, 0.5 & 0.5 \\ \hline

\textbf{Proper Subset} & Irreflexivity & Yes & 0.45 & 0.68 \\ \hline
\textbf{Proper Subset} & \makecell{Asymmetry \\ distance of \{1, 2, 3, 4\}} & Yes & 0.87, 0.96, 0.98, 0.76 & 0.57 \\ \hline
\textbf{Proper Subset} & \makecell{Average transitivity \\ \{2, 3, 4\} hops} & Yes & 0.67, 0.55, 0.33 & 0.5 \\ \hline

\end{tabular}
\caption{Results for Llama-3-8B, out-of-context learning, averaged tokens.}
\label{table:llama_out_of_context_average}
\end{table*}

\begin{table*}[h]
\centering
\small
\begin{tabular}{@{\extracolsep{4pt}}|c|c|c|c|c|}
\hline
\textbf{Relation} & \textbf{Property} & \textbf{Other properties are given} & \textbf{Accuracy} & \textbf{Baseline} \\ \hline

\Xhline{2\arrayrulewidth}

\textbf{\makecell{Strict \\ Total order}} & Irreflexivity & No & 0.19 & 0 \\  \hline
\textbf{\makecell{Strict \\ Total order}} & \makecell{Asymmetry} & No & 0.14 & 0 \\ \hline
\textbf{\makecell{Strict \\ Total order}} & \makecell{Transitivity \\ \{2, 3, 4, 5, 6, 7, 8, 9\} hops} & No & \makecell{0.87, 0.89, 0.82, 0.88, \\ 0.85, 0.95, 1.0, 0.84} & 1 \\ \hline

\textbf{\makecell{Strict \\ Total order}} & Irreflexivity & Yes & 0.23 & 0.5 \\  \hline
\textbf{\makecell{Strict \\ Total order}} & \makecell{Asymmetry \\ distance of \\ \{1, 2, 3, 4, 5, 6, 7, 8, 9\}} & Yes & \makecell{0.18, 0.3, 0.36, 0.49, \\ 0.51, 0.69, 0.82, 0.94, 1.0} & 0.18 \\ \hline
\textbf{\makecell{Strict \\ Total order}} & \makecell{Transitivity \\ \{2, 3, 4, 5, 6, 7, 8, 9\} hops} & Yes & \makecell{0.58, 0.56, 0.63, 0.52, \\ 0.5, 0.63, 0.64, 1.0} & 0.5 \\ \hline

\Xhline{2\arrayrulewidth}

\textbf{Equality} & Reflexivity & No & 0.82 & 0.89 \\  \hline
\textbf{Equality} & Average symmetry & No & 0.47 & 0.5 \\  \hline
\textbf{Equality} & \makecell{Average transitivity \\ \{2, 3, 4\} hops} & No & 0.72, 0.72, 0.56 & 0.5 \\ \hline
\textbf{Equality} & Reflexivity & Yes & 0.16 & 0.44 \\  \hline
\textbf{Equality} & \makecell{Average symmetry \\ \{1, 2, 3, 4\} hops} & Yes & 0.49, 0.53, 0.48, 0.35 & 0.5 \\ \hline
\textbf{Equality} & \makecell{Average transitivity \\ \{2, 3, 4\} hops} & Yes & 0.5, 0.5, 0.5 & 0.5 \\ \hline

\Xhline{2\arrayrulewidth}

\textbf{Proper subset} & Irreflexivity & No & 0.38 & 0.05 \\ \hline
\textbf{Proper subset} & \makecell{Asymmetry} & No & 0.36 & 0.05 \\ \hline
\textbf{Proper subset} & \makecell{Average transitivity \\ \{2, 3, 4\} hops} & No & 0.61, 0.36, 0.38 & 0.5 \\ \hline

\textbf{Proper subset} & Irreflexivity & Yes & 0.67 & 0.68 \\ \hline
\textbf{Proper subset} & \makecell{Asymmetry \\ distance of \{1, 2, 3, 4\}} & Yes & 0.9, 0.98, 1.0, 1.0 & 0.57 \\ \hline
\textbf{Proper subset} & \makecell{Average transitivity \\ \{2, 3, 4\} hops} & Yes & 0.59, 0.63, 0.66 & 0.5 \\ \hline

\end{tabular}
\caption{Results for Mistral-7B-v0.3, out-of-context learning, averaged tokens.}
\label{table:mistral_out_of_context_average}
\end{table*}

\begin{table*}[h]
\centering
\small
\begin{tabular}{@{\extracolsep{4pt}}|c|c|c|c|c|}
\hline
\textbf{Relation} & \textbf{Property} & \textbf{Other properties are given} & \textbf{Accuracy} & \textbf{Baseline} \\ \hline

\Xhline{2\arrayrulewidth}

\textbf{\makecell{Strict \\ Total order}} & Irreflexivity & No & 0.34 & 0 \\  \hline
\textbf{\makecell{Strict \\ Total order}} & \makecell{Asymmetry} & No & 0.3 & 0 \\ \hline
\textbf{\makecell{Strict \\ Total order}} & \makecell{Transitivity \\ \{2, 3, 4, 5, 6, 7, 8, 9\} hops} & No & \makecell{0.68, 0.66, 0.68, 0.64, \\ 0.63, 0.71, 0.66, 0.63} & 1 \\ \hline

\textbf{\makecell{Strict \\ Total order}} & Irreflexivity & Yes & 0.12 & 0.5 \\  \hline
\textbf{\makecell{Strict \\ Total order}} & \makecell{Asymmetry \\ distance of \\ \{1, 2, 3, 4, 5, 6, 7, 8, 9\}} & Yes & \makecell{0.38, 0.38, 0.4, 0.49, \\ 0.51, 0.43, 0.48, 0.41, 0.45} & 0.18 \\ \hline
\textbf{\makecell{Strict \\ Total order}} & \makecell{Transitivity \\ \{2, 3, 4, 5, 6, 7, 8, 9\} hops} & Yes & \makecell{0.52, 0.56, 0.53, 0.57, \\ 0.51, 0.6, 0.61, 0.69} & 0.5 \\ \hline

\Xhline{2\arrayrulewidth}

\textbf{Equality} & Reflexivity & No & 0.56 & 0.89 \\  \hline
\textbf{Equality} & Average symmetry & No & 0.51 & 0.5 \\  \hline
\textbf{Equality} & \makecell{Average transitivity \\ \{2, 3, 4\} hops} & No & 0.49, 0.52, 0.54 & 0.5 \\ \hline
\textbf{Equality} & Reflexivity & Yes & 0.48 & 0.44 \\  \hline
\textbf{Equality} & \makecell{Average symmetry \\ \{1, 2, 3, 4\} hops} & Yes & 0.36, 0.4, 0.38, 0.38 & 0.5 \\ \hline
\textbf{Equality} & \makecell{Average transitivity \\ \{2, 3, 4\} hops} & Yes & 0.57, 0.54, 0.52 & 0.5 \\ \hline

\Xhline{2\arrayrulewidth}

\textbf{Proper subset} & Irreflexivity & No & 0.46 & 0.05 \\ \hline
\textbf{Proper subset} & \makecell{Asymmetry} & No & 0.45 & 0.05 \\ \hline
\textbf{Proper subset} & \makecell{Average transitivity \\ \{2, 3, 4\} hops} & No & 0.49, 0.51, 0.48 & 0.5 \\ \hline

\textbf{Proper subset} & Irreflexivity & Yes & 0.55 & 0.68 \\ \hline
\textbf{Proper subset} & \makecell{Asymmetry \\ distance of \{1, 2, 3, 4\}} & Yes & 0.63, 0.67, 0.62, 0.67 & 0.57 \\ \hline
\textbf{Proper subset} & \makecell{Average transitivity \\ \{2, 3, 4\} hops} & Yes & 0.52, 0.51, 0.5 & 0.5 \\ \hline

\end{tabular}
\caption{Results for Llama-3.2-1B, out-of-context learning, averaged tokens.}
\label{table:llama_1b_out_of_context_average}
\end{table*}

\clearpage

\subsection{Additional Figures}

\begin{figure}[h]
    \centering
    \includegraphics[width=0.45\textwidth]{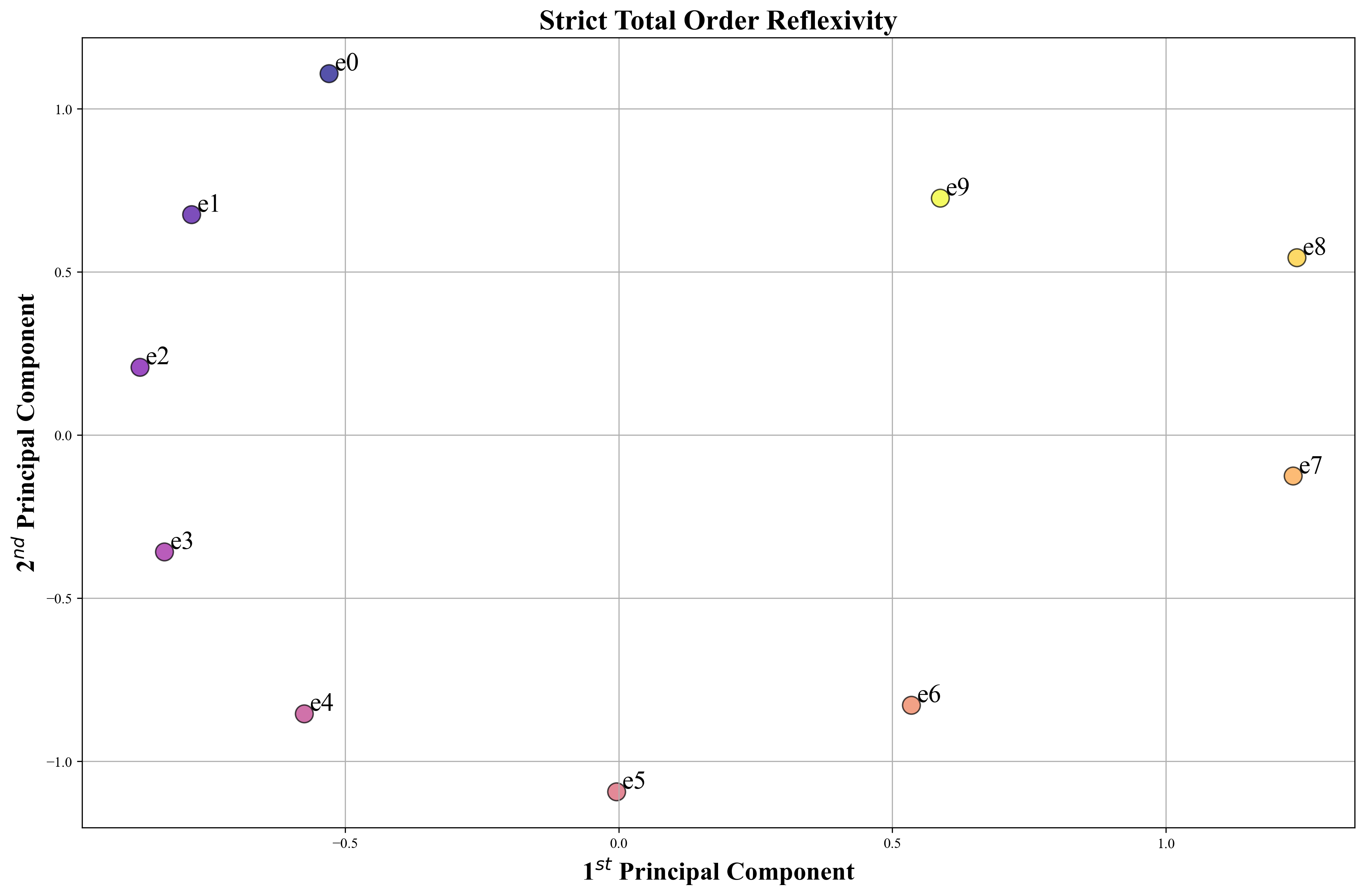}
    \caption{Mistral-7B-v0.3 strict total order, where asymmetry and transitivity are given.}
    \label{fig:mistral_total_order}
\end{figure}

\begin{figure}[h]
    \centering
    \includegraphics[width=0.45\textwidth]{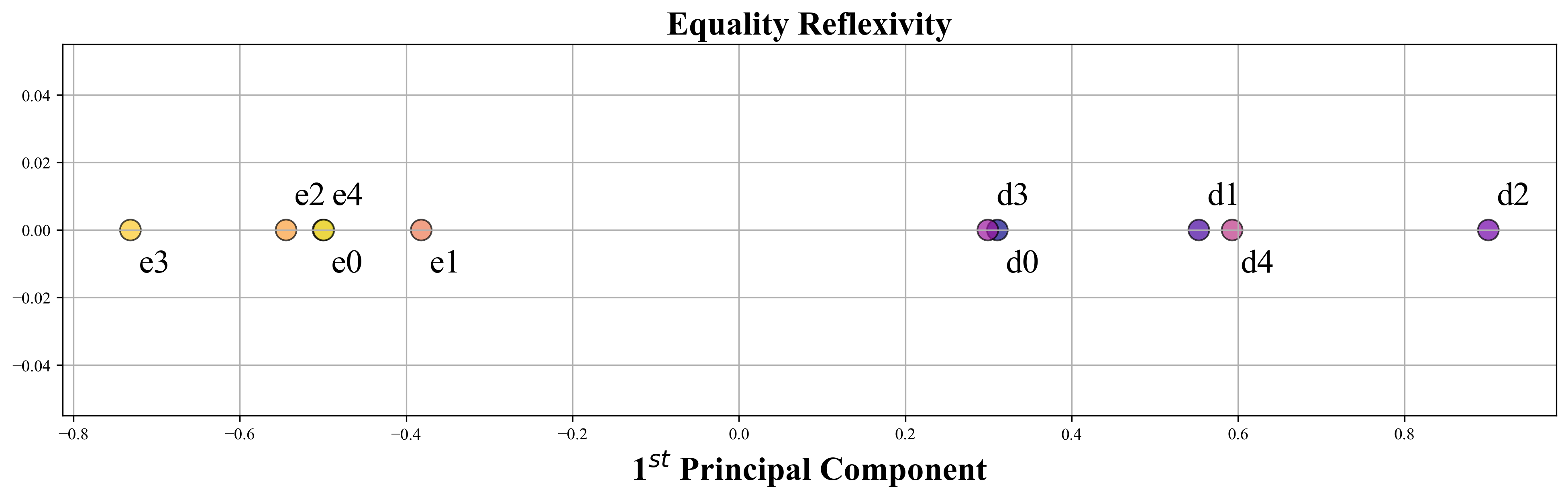}
    \caption{Mistral-7B-v0.3, equality, where symmetry and transitivity are given.}\label{fig:mistral_equality}
\end{figure}

\begin{figure}[h]
    \centering
    \includegraphics[width=0.45\textwidth]{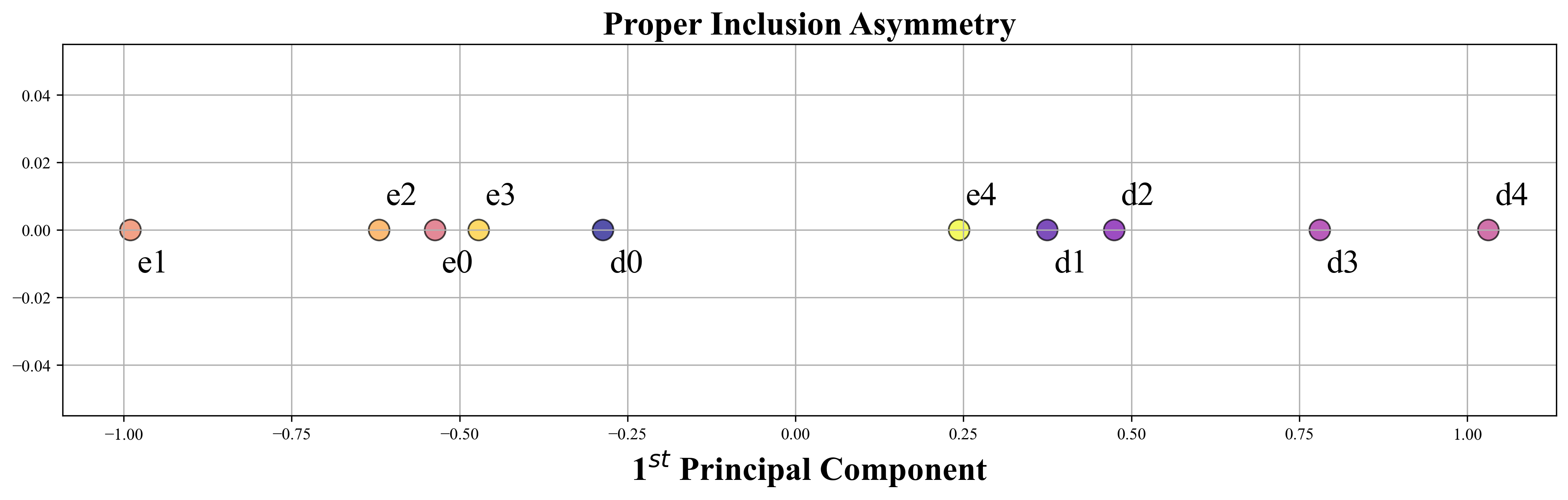}
    \caption{Mistral-7B-v0.3 proper subset, where irreflexivity and transitivity are given.}\label{fig:mistral_subset}
\end{figure}

\begin{figure}[h]
    \centering
    \includegraphics[width=0.5\textwidth]{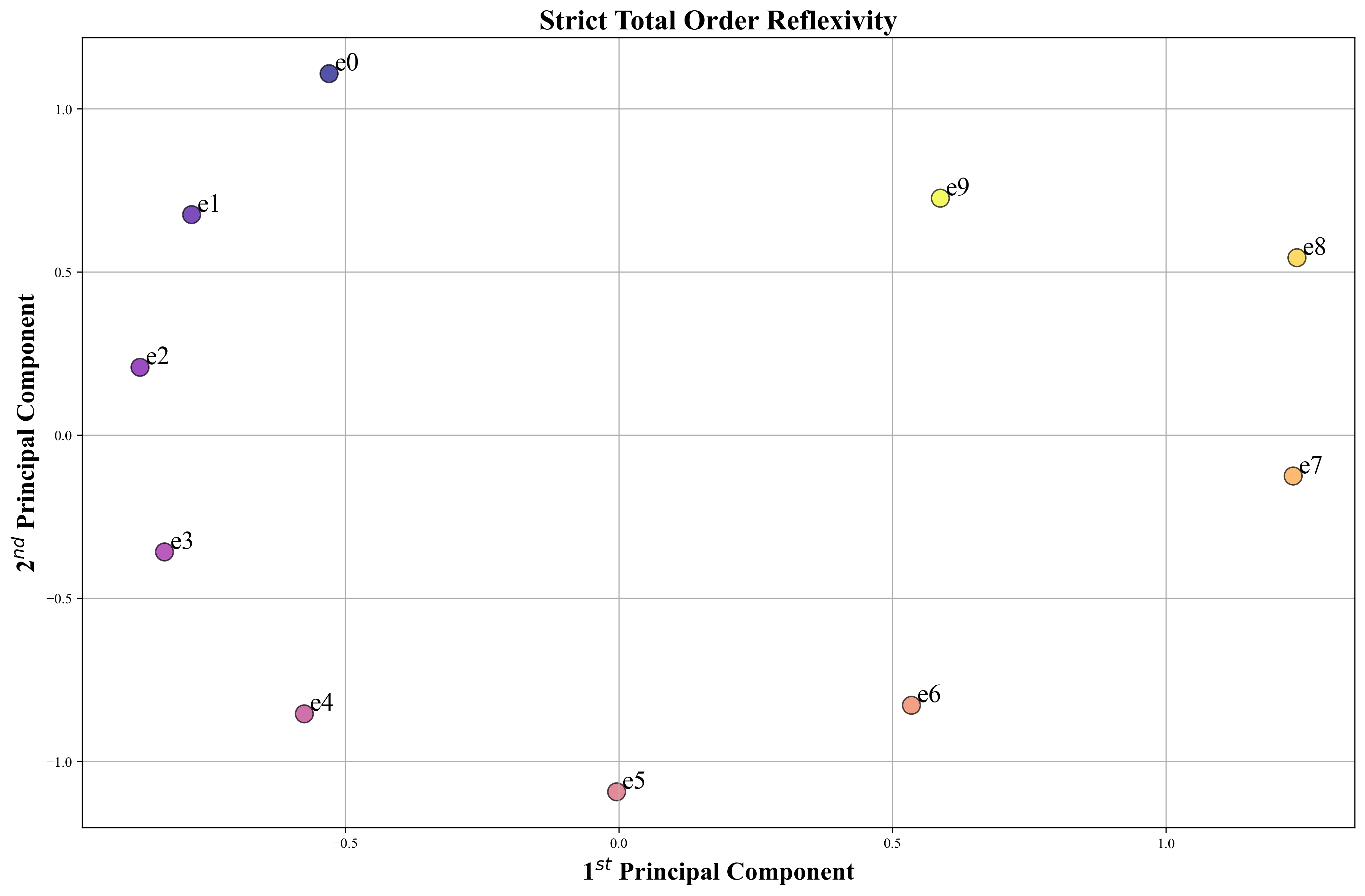}
    \caption{Llama-3.2-1B strict total order, where asymmetry and transitivity are given.}
    \label{fig:llama_1b_total_order}
\end{figure}

\begin{figure}[h]
    \centering
    \includegraphics[width=0.5\textwidth]{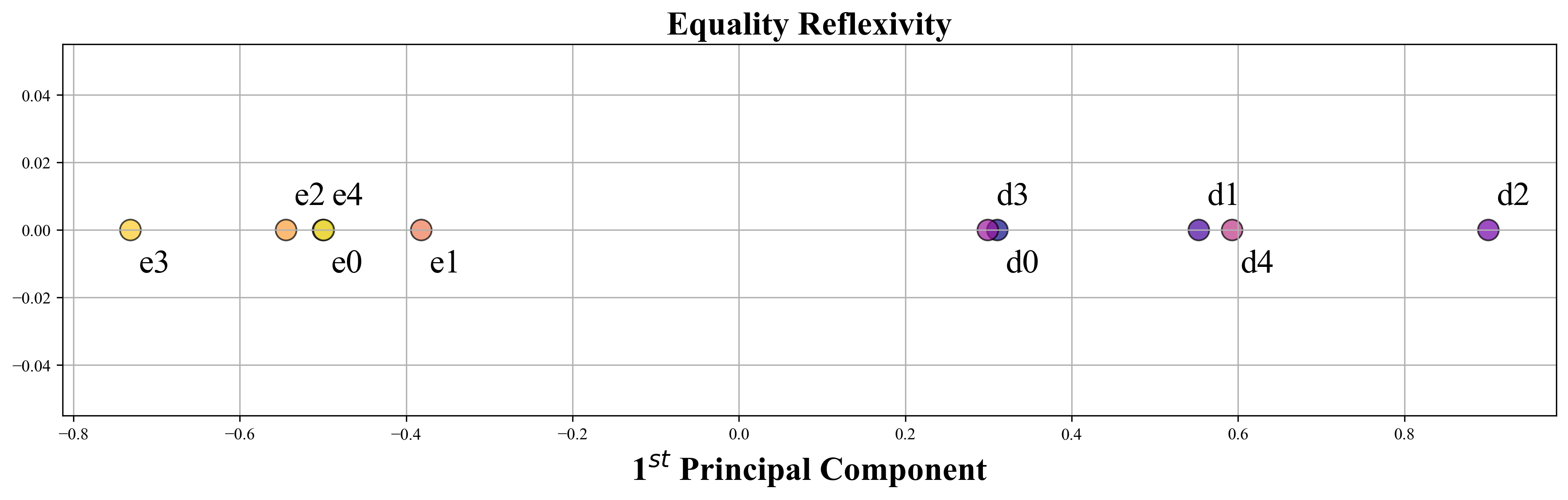}
    \caption{Llama-3.2-1B, equality, where symmetry and transitivity are given.}\label{fig:llama_1b_equality}
\end{figure}

\begin{figure}[h]
    \centering
    \includegraphics[width=0.5\textwidth]{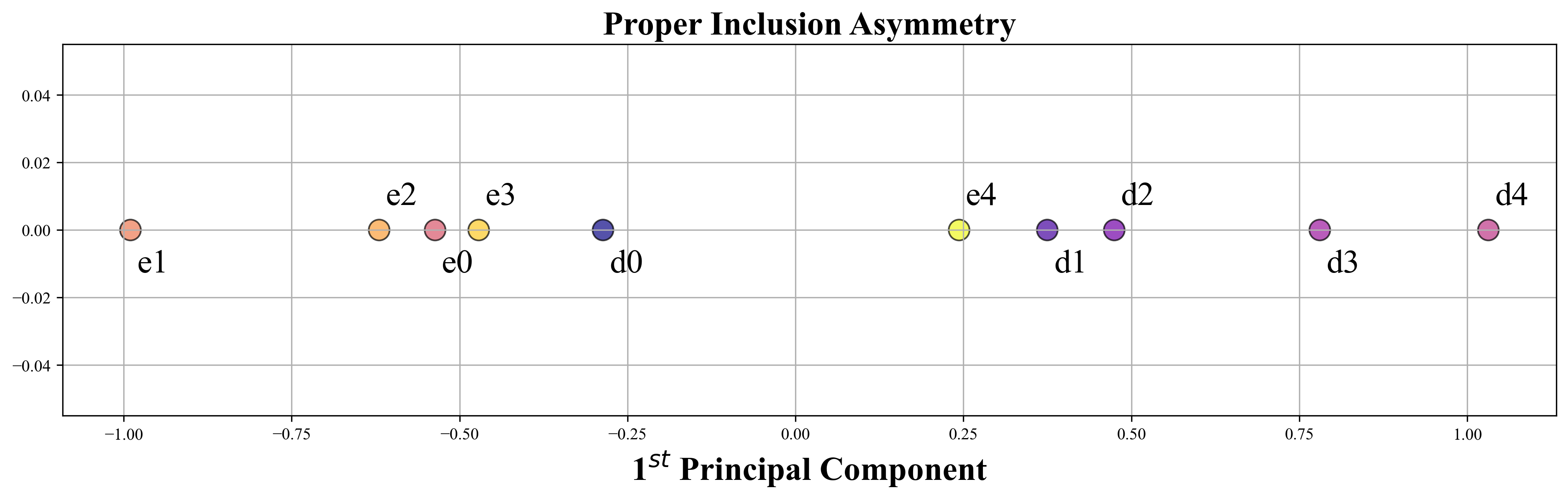}
    \caption{Llama-3.2-1B proper subset, where irreflexivity and transitivity are given.}\label{fig:llama_1b_subset}
\end{figure}

\end{document}